\documentclass[twocolumn]{article} % Method A for two-column formatting

\usepackage{preprint}
\usepackage{glossaries}
\usepackage{graphicx}

\usepackage{amsmath, amsthm, amssymb, amsfonts}

\usepackage{algorithm}
\usepackage{algpseudocode}

\usepackage[numbers,square]{natbib}
\usepackage[utf8]{inputenc}	% allow utf-8 input
\usepackage[T1]{fontenc}	% use 8-bit T1 fonts
\usepackage{xcolor}		% colors for hyperlinks
\usepackage[colorlinks = true,
            linkcolor = purple,
            urlcolor  = blue,
            citecolor = cyan,
            anchorcolor = black]{hyperref}	% Color links to references, figures, etc.
\usepackage{booktabs} 		% professional-quality tables
\usepackage{multirow}		% multi-row cells in tables
\usepackage{nicefrac}		% compact symbols for 1/2, etc.
\usepackage{microtype}		% microtypography
\usepackage{lineno}		% Line numbers
\usepackage{float}			% Allows for figures within multicol

\usepackage{lipsum}		%  Filler text

\usepackage{newfloat}
\DeclareFloatingEnvironment[name={Supplementary Figure}]{suppfigure}
\usepackage{sidecap}
\sidecaptionvpos{figure}{c}

\usepackage{titlesec}
\titlespacing\section{0pt}{12pt plus 3pt minus 3pt}{1pt plus 1pt minus 1pt}
\titlespacing\subsection{0pt}{10pt plus 3pt minus 3pt}{1pt plus 1pt minus 1pt}
\titlespacing\subsubsection{0pt}{8pt plus 3pt minus 3pt}{1pt plus 1pt minus 1pt}

\usepackage{tikz,xcolor,hyperref}

\definecolor{lime}{HTML}{A6CE39}
\DeclareRobustCommand{\orcidicon}{
	\begin{tikzpicture}
	\draw[lime, fill=lime] (0,0) 
	circle [radius=0.16] 
	node[white] {{\fontfamily{qag}\selectfont \tiny ID}};
	\draw[white, fill=white] (-0.0625,0.095) 
	circle [radius=0.007];
	\end{tikzpicture}
	\hspace{-2mm}
}
\foreach \x in {A, ..., Z}{\expandafter\xdef\csname orcid\x\endcsname{\noexpand\href{https://orcid.org/\csname orcidauthor\x\endcsname}
			{\noexpand\orcidicon}}
}
\newacronym{tsp}{TSP}{Traveling Salesman Problem}
\newacronym{tatsp}{TA-TSP}{Trigger Arc TSP}
\newacronym{grasp}{GRASP}{Greedy Randomized Adaptive Search Procedures}
\newacronym{mip}{MIP}{Mixed-Integer Programming}
\newacronym{rgc}{RGC}{Randomized Greedy Construction heuristic} 

\date{August 2, 2025}
\title{A Fast GRASP Metaheuristic for the Trigger Arc TSP with MIP-Based Construction and Multi-Neighborhood Local Search}
\shorttitle{A Fast GRASP Metaheuristic for the Trigger Arc TSP}

\usepackage{authblk}

\author[1,2]{Joan Salvà Soler*\orcidA{}} 
\author[1]{Grégoire de Lambertye\orcidB{}}

\affil[1]{Technische Universität Wien}
\affil[2]{H2O.ai}

\begin{document}

\twocolumn[ % Method A for two-column formatting
  
\maketitle

\begin{abstract}
The Trigger Arc Traveling Salesman Problem (TA-TSP) extends the classical TSP by introducing dynamic arc costs that change when specific \textit{trigger} arcs are traversed, modeling scenarios such as warehouse operations with compactable storage systems. This paper introduces a GRASP-based metaheuristic that combines multiple construction heuristics with a multi-neighborhood local search. The construction phase uses mixed-integer programming (MIP) techniques to transform the TA-TSP into a sequence of tailored TSP instances, while the improvement phase applies 2-Opt, Swap, and Relocate operators. Computational experiments on MESS 2024 competition instances achieved average optimality gaps of 0.77\% and 0.40\% relative to the best-known solutions within a 60-second limit. On smaller, synthetically generated datasets, the method produced solutions 11.3\% better than the Gurobi solver under the same time constraints.
The algorithm finished in the top three at MESS 2024, demonstrating its suitability for real-time routing applications with state-dependent travel costs.
\end{abstract} 

 % (optional)

\vspace{0.35cm}

] % Method A for two-column formatting

%\begin{multicols}{2} % Method B for two-column formatting (doesn't play well with line numbers), comment out if using method A

%%%%%%%%%%%%%%%  Main text   %%%%%%%%%%%%%%%
% \linenumbers

\section{Introduction}

The \gls{tsp} is a foundational problem in combinatorial optimization~\cite{Applegate2011}. 
However, its traditional formulation assumes that travel costs between cities are static, which is insufficient for many real-world applications where the cost of traversing a path can change based on prior actions or the system's current state.
An example of this occurs in warehouses that use mobile shelving units. Accessing an item may require moving a shelf, which can block other aisles or alter the time and cost needed to reach them. This path-dependent cost structure necessitates a more sophisticated modeling approach than the classical TSP can provide.

To address these behaviors, \citet{Cerrone} introduced the \gls{tatsp}, which is a variant of the \gls{tsp} where traversing a \textit{trigger} arc can dynamically alter the cost of a \textit{target} arc later in the tour. 
The problem defines specific relationships between pairs of trigger and target arcs. 
When a trigger is traversed, it modifies the cost of its associated target. 
If multiple triggers for the same target are visited, only the last one encountered before the target takes effect. 
This novel formulation provides a powerful tool for modeling problems with state-dependent costs.

This paper presents a GRASP-based approach to the \gls{tatsp}, developed for the Metaheuristics Summer School (MESS) 2024 competition~\cite{mess2024}. 
A top-three position was secured by proposing a scalable and fast \gls{grasp}-based algorithm to find high-quality solutions. 
A \gls{mip}-based construction heuristic is employed to generate diverse and feasible initial solutions, followed by a local search procedure for iterative improvement.
The methodology and the results of applying the \gls{grasp} algorithm to a set of benchmark instances are outlined, demonstrating effectiveness and efficiency in tackling the \gls{tatsp}.

\section{Related Work}
\label{sec:related_work}

\subsection{Variants of the \gls{tsp}}

The \gls{tsp} is fundamentally about discovering the most efficient route that connects a given set of nodes~\cite{Applegate2011}. 
Its NP-hard nature makes exact solutions for large instances computationally infeasible, necessitating the use of heuristics and metaheuristics.
Their utility becomes even more pronounced when dealing with the intricacies of dynamic TSP variants.
These variants introduce real-world complexities, including problems on temporal graphs where edge access is time-sensitive~\cite{timedependenttsp}, 
TSPs where visiting a node has the immediate consequence of deleting specific edges~\cite{traversaldependenttsp}, and scenarios where the cost structure of a tour is directly affected by the historical sequence of visited locations~\cite{Bossek2020}.

The \gls{tatsp}, introduced by~\citet{Cerrone}, is a variant that introduces dynamic arc costs influenced by the solution's path. 
In the \gls{tatsp}, the cost of traversing a specific arc can be reset if a predefined trigger arc is traversed beforehand, 
with only the most recently encountered trigger arc affecting the target arc's cost. In their introduction paper,~\citet{Cerrone} 
propose a \gls{mip} model for the \gls{tatsp}. Their approach provides a precise mathematical formulation 
but is limited in scalability, being able to tackle instances with up to 25 nodes and 10,000 trigger-target relations. 
The problem's inherent complexity and the practical need for fast solutions for large-scale instances underscore the need for robust metaheuristic approaches.

\subsection{Metaheuristics for the TSP}
\label{sec:metaheuristics_tsp}

Metaheuristics are a class of optimization algorithms that provide approximate solutions to complex problems by efficiently exploring and exploiting a search space~\cite{Gendreau2010}.
Given the NP-hard nature of the \gls{tsp}, metaheuristics have been extensively explored to find high-quality solutions within a reasonable computational time~\cite{toaza2023}.
Prominent examples include Genetic (or Evolutionary) Algorithms~\cite{Larranaga1999}, Tabu Search~\cite{Knox1994}, and Simulated Annealing~\cite{anttsp}.
A classic local search procedure applied to the \gls{tsp} is the 2-Opt algorithm~\cite{Croes1958}.

\subsection{Greedy Randomized Adaptive Search Procedures (\gls{grasp})}

The present work is based on the \gls{grasp} metaheuristic, introduced by Feo and Resende~\cite{Feo1995}, which is a two-phase approach that combines a construction heuristic and a local search procedure.
The construction phase leverages a greedy randomized strategy to build a diverse set of feasible solutions. Following this, the local search phase systematically explores the neighborhood of each constructed solution to find a local optimum with respect to a set of neighborhood structures.
The best solution found across all iterations is returned. The two-phase structure of \gls{grasp} inherently balances diversification (exploration) and intensification (exploitation) of the search space, which is advantageous for complex fitness landscapes.
Due to its robust performance and adaptability, \gls{grasp} has been successfully applied to the classical \gls{tsp}, often achieving results that are competitive with highly specialized heuristics~\cite{Oliveira2004}. 
Its effectiveness can be further enhanced through advanced techniques. Reactive GRASP, for instance, dynamically adjusts its parameters during the search to adapt to the problem's characteristics~\cite{Prais2000}. 
Path-Relinking is an intensification strategy that explores trajectories between elite solutions to find improved ones~\cite{Resende2019}. The consistent success of \gls{grasp} and its variants in related routing problems makes it a compelling choice for addressing the \gls{tatsp}.

\section{Methodology}

\subsection{Problem Definition}

The \gls{tatsp} is formally defined on a directed graph $G = (N, A)$, where $N$ is the set of nodes and $A$ is the set of directed arcs. Node $0 \in N$ is designated as the starting node (depot). The problem incorporates the following key elements:

\begin{itemize}
\item \textbf{Base arc costs}: Let $c(a) \in \mathbb{R}^+$ for all $a \in A$ represent the base cost of traversing arc $a$.
\item \textbf{Relations}: For each arc $a_j \in A$, a set of relations $R_{a_j} = \{(a_i, a_j) \mid a_i \in A\}$ is defined, where each relation connects a trigger arc $a_i$ to the target arc $a_j$.
\item \textbf{Relation costs}: Let $c(r) \in \mathbb{R}^+$ for all $r \in R$ represent the cost associated with relation $r$ when active, where $R = \bigcup_{a \in A} R_a$ is the set of all relations.
\end{itemize}

Given a Hamiltonian cycle $T = (a_i, a_j, \ldots, a_{|N|})$ starting at node $0$, a relation $r = (a_i, a_j) \in R$ is active if and only if:
\begin{enumerate}
\item Both arcs $a_i$ and $a_j$ belong to the tour $T$.
\item Arc $a_i$ precedes arc $a_j$ in the traversal sequence.
\item There is no other relation $r' = (a_k, a_j) \in R_{a_j}$ such that $a_k$ follows $a_i$ in $T$.
\end{enumerate}

The cost of traversing an arc $a \in A$ in tour $T$ is determined as:
\begin{equation}
\text{cost}(a) = \begin{cases}
c(r) & \text{if relation } r \in R_a \text{ is active} \\
c(a) & \text{if no relation in } R_a \text{ is active}
\end{cases}
\end{equation}

The objective of the \gls{tatsp} is to find a Hamiltonian cycle $T^*$ that minimizes the total tour cost:
\begin{equation}
T^* = \arg\min_{T} \sum_{a \in T} cost(a)
\end{equation}

The \gls{tatsp} generalizes the \gls{tsp}. If the relation set within the \gls{tatsp} is empty, 
the problem formulation simplifies directly to that of the traditional \gls{tsp}.
The two problems share the same set of feasible solutions, which is the set of all Hamiltonian cycles.
However, there is no natural reduction from the \gls{tatsp} to the \gls{tsp} because of the path-dependency of the cost function.
For modeling purposes, it is useful to reformulate the cost function. Instead of viewing the cost of a relation as a replacement for the base arc cost, it can be considered as an additional cost (or discount). For each relation $r = (a_i, a)$ with an associated cost $c(r)$, a delta cost can be defined as $\Delta(r) = c(r) - c(a)$. The cost of traversing an arc $a$ in a tour $T$ is then $c(a) + \Delta(r)$ if relation $r$ is active, and $c(a)$ otherwise. This separates the base tour cost from the additional costs incurred by active relations, which is a convenient perspective for developing heuristics and mathematical models.

\subsection{A \gls{grasp} for the \gls{tatsp}}

The original version of the \gls{grasp} metaheuristic, as introduced by~\cite{Feo1995}, is leveraged in this study to solve the \gls{tatsp}. 
The algorithm, as described in Algorithm~\ref{alg:grasp_general}, consists of two main phases: a construction phase and a local search phase.
As construction heuristics, we consider the Simple Randomized Construction, the Randomized Greedy Construction, and the \gls{mip}-based Random Perturbation.
As the local search procedure, a multi-neighborhood local search with a first-improvement policy is employed, described in Algorithm~\ref{alg:local_search}.
The stopping criterion for the local search is the local optimality with respect to all the neighborhood structures considered.

\begin{algorithm}
\caption{General GRASP Framework}
\label{alg:grasp_general}
\begin{algorithmic}[1]
\State $S^* \leftarrow \emptyset$ \Comment{Best solution found}
\State $f^* \leftarrow \infty$ \Comment{Best objective value}
\For{$i = 1$ to $\text{MaxIterations}$}
    \State $S \leftarrow \text{RandomizedConstruction}()$
    \State $S \leftarrow \text{LocalSearch}(S)$
    \If{$f(S) < f^*$}
        \State $S^* \leftarrow S$
        \State $f^* \leftarrow f(S)$
    \EndIf
\EndFor
\State \Return $S^*$
\end{algorithmic}
\end{algorithm}

\begin{algorithm}
\caption{Multi-Neighborhood Local Search}
\label{alg:local_search}
\begin{algorithmic}[1]
\State \textbf{Input:} Tour $T$, Neighborhoods $\mathcal{N} = \{\mathcal{N}_1, \mathcal{N}_2, \mathcal{N}_3\}$
\State \textbf{Output:} Locally optimal tour $T'$
\Repeat
    \State $\text{improved} \leftarrow \text{false}$
    \For{each neighborhood $\mathcal{N}_i \in \mathcal{N}$}
        \For{each move $m \in \mathcal{N}_i(T)$}
            \State $T' \leftarrow \text{ApplyMove}(T, m)$
            \State $\Delta \leftarrow \text{EvaluateMoveCost}(T, m)$ \Comment{Compute cost change with trigger effects}
            \If{$\Delta < 0$}
                \State $T \leftarrow T'$
                \State $\text{improved} \leftarrow \text{true}$
                \State \textbf{break} \Comment{First improvement strategy}
            \EndIf
        \EndFor
        \If{$\text{improved}$}
            \State \textbf{break} \Comment{Restart with first neighborhood}
        \EndIf
    \EndFor
\Until{$\text{improved} = \text{false}$}
\State \Return $T$
\end{algorithmic}
\end{algorithm}

\subsection{Construction Heuristics} \label{sec:constr_heuristic}
This section defines the construction heuristics compared in the experiments.

\paragraph{Simple Randomized Construction}
This heuristic serves as a baseline method for generating feasible solutions.
It constructs a tour iteratively, starting from the depot (node 0).
At each step, it identifies all unvisited neighbors of the last node added to the tour.
From this set of feasible next nodes, one is selected uniformly at random and appended to the tour.
This process continues until all nodes have been visited.
The final node is only considered feasible if it has an arc connecting back to the depot, thus ensuring a valid Hamiltonian cycle.
This method does not consider any cost information during construction, relying purely on random selection to explore the solution space.

\paragraph{Randomized Greedy Construction}
This method is a semi-greedy heuristic, similar to the previous approach, builds a solution one node at a time.
However, the choice of the next node is guided by cost information. At each step, for every feasible successor node, the exact incremental cost of adding the corresponding arc to the partial tour is calculated. This evaluation is non-trivial, as it must account for any changes in the total tour cost due to the activation of trigger-target arc relations.
A Restricted Candidate List (RCL) is then formed, containing the most promising arcs based on their incremental cost. The size of the RCL is controlled by a randomization parameter, $\alpha \in [0,1]$, which represents the share of neighbors added to the list. An arc is then selected uniformly at random from the RCL, and the corresponding node is added to the tour. A value of $\alpha=0$ results in a purely greedy construction where the best arc is always chosen, while $\alpha=1$ allows for a random selection from all feasible arcs.
  
\paragraph{\textit{\gls{mip}-based Random Perturbation}}
Inspired by the literature on metaheuristic hybridization~\cite{Blum2016}, a novel construction heuristic transforms the \gls{tatsp} into a classic \gls{tsp}, thereby ignoring all trigger-target relations.
This is partially motivated by the fact that the \gls{tsp} associated with the \gls{tatsp} is generally easy to solve with a \gls{mip} solver, which can quickly find primal solutions due to the effective \gls{tsp}-specific primal heuristics available in many solvers.

The core idea is to perturb the original edge costs, $c_{ij}$, to create a new \gls{tsp} instance.
This instance is then solved to optimality or near-optimality. The resulting tour, or a pool of high-quality tours, is then evaluated using the true \gls{tatsp} objective function.

Two perturbation methods are explored. In the \textit{Additive Perturbation} approach, costs are modified using an additive noise factor controlled by a parameter $\alpha$. The new cost, $c'_{ij}$, is calculated as $c'_{ij} = c_{ij} + \alpha \cdot U(-1, 1)$, where $U(-1, 1)$ is a random variable from a uniform distribution between $-1$ and $1$. Alternatively, the \textit{Multiplicative Perturbation} method scales the costs by a random factor controlled by a parameter $\beta$, resulting in $c'_{ij} = c_{ij} \cdot (\beta \cdot U(0, 1))$.

\paragraph{\textit{\gls{mip}-based Biased Perturbation}}
This method attempts to incorporate relational information into the \gls{tsp} model by perturbing arc costs based on an estimated probability of relation activation. The process begins by generating a random permutation of nodes, $P$, which serves as a prior for estimating arc usage. The probability of an arc $(i, j)$ being used is assumed to be inversely proportional to the cyclic distance between nodes $i$ and $j$ in the prior permutation $P$. This is defined as $p_{ij} = 1/d_{ij}$, where $d_{ij} = \min(|pos(i)-pos(j)|, |N| - |pos(i)-pos(j)|)$. For a relation $r$ from a trigger arc $a=(a_1, a_2)$ to a target arc $b=(b_1, b_2)$, the activation probability $p_r$ is estimated by considering the usage probabilities of both arcs and their proximity. It is defined as $p_r = p_{a_1a_2} \cdot p_{b_1b_2} / d_{a_2b_1}^\beta$, where $d_{a_2b_1}$ is the cyclic distance between the trigger's endpoint and the target's startpoint, and $\beta$ is a tunable parameter. The costs of the arcs in the base \gls{tsp} instance are modified to reflect the expected impact of the relations. For each relation $r$ between trigger arc $b$ and target arc $a$, a penalty proportional to the relation's activation probability is added to the costs of both arcs. The new cost of an arc $k \in \{a,b\}$ is updated as $c'(k) = c(k) + \alpha \cdot p_r \cdot c_r$, where $c_r$ is the cost when relation $r$ is active, and $\alpha$ is a parameter controlling the penalty's magnitude. Finally, the modified \gls{tsp} instance is solved, and the resulting tours are evaluated on the original \gls{tatsp} instance to find the best solution.

\subsection{Local search}
The local search phase is designed to iteratively improve a solution by exploring its neighborhood. A multi-neighborhood variable search procedure systematically applies a set of local search operators. The search follows a first-improvement strategy: as soon as a move that reduces the total tour cost is found, the move is applied, and the search is restarted from the beginning of the neighborhood list.
This process continues until no further improvements can be found in any of the defined neighborhoods, at which point the solution is considered locally optimal. The neighborhood structures used are 2-Opt, Swap, and Relocate, with descriptions provided hereafter.

\paragraph{2-Opt} The 2-Opt move is a powerful neighborhood operator for path-improvement~\cite{Croes1958}. Given a tour, it works by removing two non-adjacent arcs, $(v_i, v_{i+1})$ and $(v_j, v_{j+1})$, and reconnecting the two resulting paths in the only other possible way. This is equivalent to reconnecting them with arcs $(v_i, v_j)$ and $(v_{i+1}, v_{j+1})$, which has the effect of reversing the segment of the tour between nodes $v_{i+1}$ and $v_j$.

\paragraph{Swap} The Swap (or Exchange) move consists of exchanging the positions of two nodes, $v_i$ and $v_j$, in the tour. The depot node is not considered for this operation. This move alters four arcs in the tour: the two arcs entering and the two arcs leaving the swapped nodes.

\paragraph{Relocate} The Relocate (or Insertion) move selects a node $v_i$ and removes it from its current position in the tour. It is then re-inserted into a different position, between two adjacent nodes:
the tour $(\ldots, v_i, \ldots, v_j, \ldots)$ becomes $(\ldots, v_{i-1}, v_{i+1}, \ldots, v_j, v_i, v_{j+1}, \ldots)$.

\subsection{Delta Evaluation}
A critical aspect of any iterative improvement algorithm is the efficient evaluation of moves, often achieved through fast delta evaluation to avoid recomputing the entire solution cost.

For sequential construction heuristics, such as the Randomized Greedy method, delta evaluation is successfully employed. When a new node is added to a partial tour, only the cost of the new arc and its trigger effects from existing arcs need to be computed. This is feasible because the tour is extended in a single direction, so the trigger states of existing arcs remain unchanged.

However, this approach is not practical for the local search phase. Standard neighborhood moves, like 2-Opt or Swap, can drastically alter the tour's sequence. A single change can trigger a cascade effect, leading to non-local cost modifications across the entire tour. Implementing true delta evaluation would require a complex procedure to track every affected arc, which would negate its performance benefits. Consequently, the local search implementation re-evaluates the entire tour cost after every move to ensure correctness and simplicity.

\section{Computational Experiments}
This section presents the computational experiments conducted to evaluate the performance of the proposed \gls{grasp} algorithm.

\subsection{Datasets}

Three datasets were used to evaluate the \gls{grasp} algorithm and assess its performance across problems having different characteristics.

The first dataset contains 180 synthetic instances generated as described in~\citet{Cerrone}.
Each instance is generated using a random generator that creates complete directed graphs $G(N,A)$ 
with nodes placed within a 5 km square Euclidean space. 
For each pair of nodes $i, j \in N$, directed arcs $(i, j)$ and $(j, i)$ are created and associated with a base cost $c(a)$ 
corresponding to the Euclidean distance between the nodes. The synthetic instances are divided into three scenarios 
(Balanced, Increase, Decrease), with 60 instances per scenario. Each scenario includes instances with $n \in \{10, 15, 20, 25\}$ 
nodes and $\lfloor n/2 \rfloor, 2n, 4n, 8n, 16n$ relations per graph. For each combination of node count and relation count, three 
different instances are created.
Relations $r = (a_i, a_j) \in R$ are generated by randomly selecting pairs of arcs $a_i, a_j \in A$. The cost of each relation $c(r)$ 
is determined by generating a random value in the range $[c(a_j)/2, 2c(a_j)]$. For brevity, this dataset will hereafter be referred to as RG.

Three scenarios are characterized as follows:
\begin{itemize}
    \item \textbf{Balanced}: Relations have random costs that can be either greater or less than the base arc cost, in the range $[c(a_j)/2, 2c(a_j)]$.
    \item \textbf{Increase}: Relations have random costs greater than the base arc cost, in the range $[c(a_j), 2c(a_j)]$.
    \item \textbf{Decrease}: Relations have random costs less than the base arc cost, in the range $[c(a_j)/2, c(a_j)]$.
\end{itemize}

Two additional datasets, provided by the competition organizers, offer realistic scenarios with varying problem sizes and complexities. These will be referred to as C1 and C2. The first of these contains 21 instances, with an average of 580 arcs and 47,834 relations. The second, larger dataset has 34 instances, averaging 1,267 arcs and 556,903 relations. Table~\ref{tab:dataset_characteristics} summarizes the key characteristics of all three datasets.

\begin{table*}[t]
    \setlength{\belowcaptionskip}{8pt}
    \caption{Dataset Characteristics}
    \label{tab:dataset_characteristics}
    \centering
    \begin{tabular}{lrrrrrr}
        \toprule
        \textbf{Set} & \textbf{\# Inst.} & \textbf{\# Nodes} & \textbf{\# Arcs} & \textbf{\# Rels} & \textbf{Avg Arc Cost} & \textbf{Avg Rel Cost} \\
        & & \textbf{(Min/Avg/Max)} & \textbf{(Min/Avg/Max)} & \textbf{(Min/Avg/Max)} & & \\
        \midrule
        RG & 180 & 10 / 18 / 25 & 90 / 320 / 600 & 100 / 2092 / 10000 & $2587.19$ & $3042.15$ \\
        C1 & 21 & 20 / 40 / 60 & 200 / 1267 / 2700 & 1732 / 556903 / 4527944 & $6.29$ & $6.30$ \\
        C2 & 34 & 18 / 68 / 142 & 90 / 580 / 1562 & 1144 / 47834 / 208020 & $1.00$ & $1.08$ \\
        \bottomrule
    \end{tabular}
\end{table*}

\subsection{Experimental Setup}

The \gls{grasp} algorithm is implemented in C++20, and uses the Boost library for additional data structures.
The Gurobi optimization suite (Gurobi Optimizer version 12.0.2 build v12.0.2rc0) is used for solving the \gls{mip} instances arising from the \gls{mip}-based heuristics.

All computational experiments were conducted on a machine equipped with an Apple M1 Pro processor running macOS 24.5.0 and an ARM64 architecture with 16GB of RAM.
The complete implementation, along with all instances used in the experiments, is publicly available at \url{https://github.com/jsalvasoler/trigger_arc_tsp}.

\subsection{Metrics}

To evaluate the performance of the \gls{grasp} algorithm, the optimality gap is reported, which measures how close the solutions are to the best possible solutions. The gap is calculated as:

\begin{equation}
\text{Gap} = \frac{\text{Solution Cost} - \text{Best Known Cost}}{\text{Best Known Cost}} \times 100\%
\end{equation}

For the competition instances (C1 and C2), the best solutions achieved during the competition are used as the reference point. Section~\ref{sec:grasp_settings_for_competition} provides more details on the settings used in the competition.
For the synthetic instances (RG), the best integer solution obtained after running the Gurobi \gls{mip} solver with a time limit of 1 minute and the default settings is used.
For the randomized construction heuristics, the success rate is also measured, i.e., the percentage of instances for which a solution was found.

\subsection{Benchmarking the MIP model for the \gls{tatsp}}

The Gurobi solver~\cite{gurobi} is used to solve the \gls{mip} model defined in Appendix~\ref{sec:mip_model}.
The competition instances are significantly larger than the instances in the RG dataset. In the context of \gls{mip} modeling, this is reflected in the fact
that Gurobi can only solve 6 / 21 instances of the C1 dataset, 1 / 34 instances of the C2 dataset, but every instance of the RG dataset.
By solving, it is meant that the solver was able to write the model in a reasonable time frame, and was able to provide a solution with a gap less than 1\% in under 5 hours.
In many cases, the solver was running out of CPU memory.
Therefore, only the Gurobi results for the RG dataset are reported, in which a time limit of 1 minute and the default settings are used.

\begin{table}[h]
    \caption{Performance of Gurobi \gls{mip} solver on the RG dataset of synthetic instances. Results show average computation time and mean gap (\%) across different scenarios and problem sizes.}
    \label{tab:gurobi_performance}
    \centering
    \begin{tabular}{lrrrr}
        \toprule
        \textbf{Scenario} & \textbf{Nodes} & \textbf{Time (s)} & \textbf{Gap (\%)} \\
        \midrule
        \multirow{4}{*}{Balanced} & 10 & $1.89 \pm 2.81$ & $0.0 \pm 0.0$ \\
        & 15 & $24.04 \pm 28.83$ & $34.3 \pm 60.4$ \\
        & 20 & $34.95 \pm 29.40$ & $57.9 \pm 76.6$ \\
        & 25 & $40.84 \pm 24.70$ & $242.4 \pm 416.8$ \\
        \midrule
        \multirow{4}{*}{Decrease} & 10 & $3.90 \pm 7.38$ & $0.0 \pm 0.0$ \\
        & 15 & $25.13 \pm 27.40$ & $41.6 \pm 77.4$ \\
        & 20 & $40.14 \pm 26.30$ & $102.9 \pm 146.2$ \\
        & 25 & $48.38 \pm 19.54$ & $252.9 \pm 300.2$ \\
        \midrule
        \multirow{4}{*}{Increase} & 10 & $3.37 \pm 7.21$ & $0.0 \pm 0.0$ \\
        & 15 & $22.99 \pm 27.08$ & $36.8 \pm 83.6$ \\
        & 20 & $36.26 \pm 30.05$ & $97.1 \pm 161.7$ \\
        & 25 & $41.51 \pm 25.37$ & $154.8 \pm 215.0$ \\
        \bottomrule
    \end{tabular}
\end{table}

The three scenarios show similar behavior: all small instances (10 nodes) are solved to optimality, while both computation time and mean gap increase with problem size.
The Decrease scenario exhibits slightly higher gaps and longer runtimes compared to the other scenarios.

\subsection{Randomized Greedy Construction heuristic}
\begin{figure}[h]
    \centering
    \includegraphics[width=0.49\textwidth]{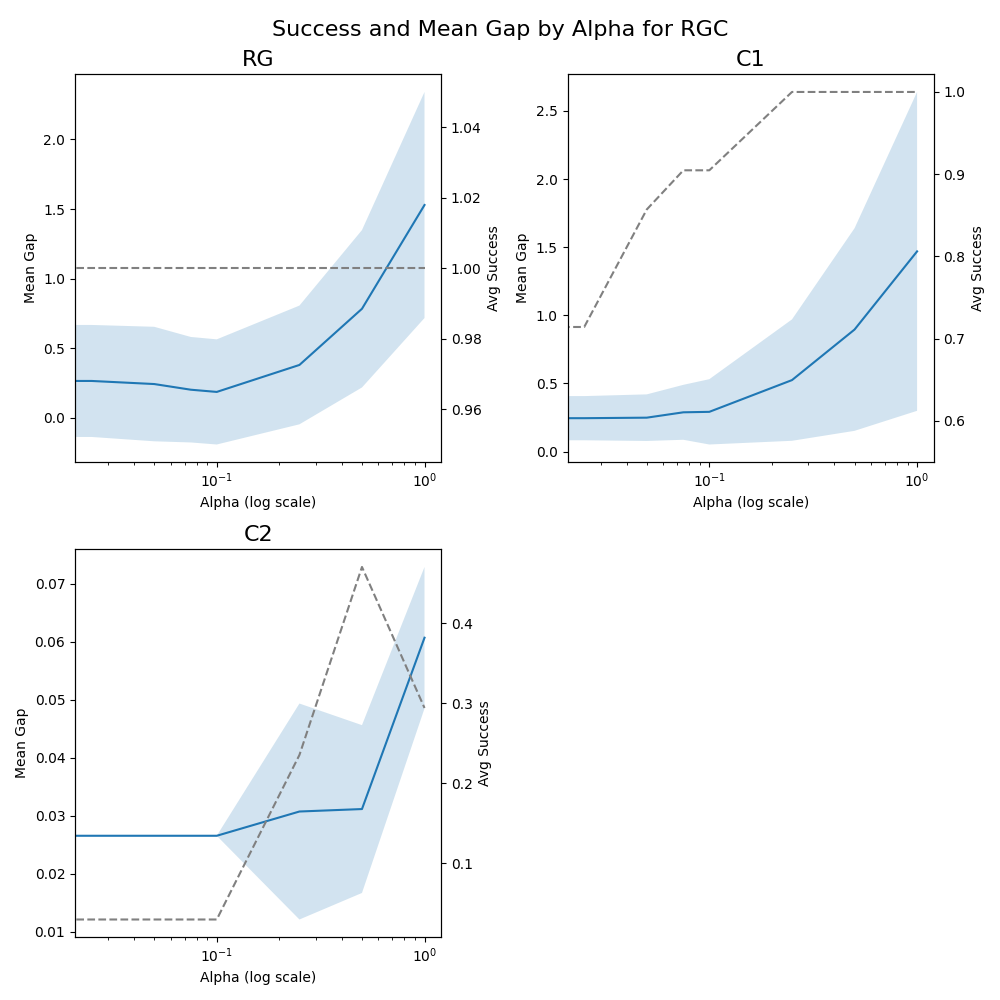}
    \caption{Performance comparison of RGC on the three datasets. Each plot shows average gap (\%) and success rate across different values of $\alpha$ for a specific dataset.}
    \label{fig:alpha_vs_mean_gap_randomized_greedy}
\end{figure}

The effect of the parameter $\alpha$ on the performance of the RGC is investigated by evaluating a grid of $\alpha$ values within the interval $[0, 1]$. 
For each value of $\alpha$ in this grid, the RGC is run for 10 trials on each instance, and the best solution found across the trials is recorded. 
Recall that $\alpha \in [0, 1]$ controls the restricted candidate list size, i.e., the fraction of feasible edges considered in the greedy step, sorted by insertion cost.
Figure~\ref{fig:alpha_vs_mean_gap_randomized_greedy} shows that increasing values of~$\alpha$ result in larger gaps (as edge insertions become, on average, more expensive) and higher success rates (as more edges are considered). For dataset~C2, some variability in the success rate is observed; however, this effect is expected to diminish with a larger number of trials. The results suggest that $\alpha = 0.1$ provides a favorable trade-off between gap and success rate, and this value is adopted for the remainder of the experiments.

\subsection{MIP-based Randomized Construction}

Both additive~($\alpha$) and multiplicative~($\beta$) perturbation strategies are evaluated for the MIP-based random perturbation construction heuristic across different parameter values. The time limit of the MIP solver is set to 2~seconds.
Recall that solving the \gls{mip} problems to optimality is not relevant in this context, since the objective that the \gls{mip} optimizes is not
the same as the one of the \gls{tatsp}. In the context of a \gls{grasp}, exploring more initial solutions and applying local search to them is more advantageous.

\begin{table}[h]
    \caption{Performance of \gls{mip} Random Perturbation with additive noise ($\alpha$). Results show the average gap (\%) across different datasets and parameter values.}
    \label{tab:mip_alpha_results}
    \centering
    \setlength{\tabcolsep}{4pt}
    \begin{tabular}{lcccc}
        \toprule
        \textbf{Set} & \boldmath{$\alpha = 0.0$} & \boldmath{$\alpha = 0.1$} & \boldmath{$\alpha = 1.0$} & \boldmath{$\alpha = 10.0$} \\
        \midrule
        RG & $81.7 \pm 82.7$ & $70.9 \pm 75.6$ & $70.6 \pm 76.9$ & $\mathbf{68.9 \pm 75.3}$ \\
        C1 & $5.5 \pm 4.1$ & $\mathbf{4.3 \pm 3.2}$ & $7.3 \pm 3.9$ & $59.1 \pm 36.7$ \\
        C2 & $13.2 \pm 5.0$ & $\mathbf{6.6 \pm 2.2}$ & $7.0 \pm 2.4$ & $7.3 \pm 2.7$ \\
        \bottomrule
    \end{tabular}
\end{table}

The results for the additive perturbation parameter $\alpha$ show that moderate values ($\alpha = 0.1$) generally perform best across all datasets. On the RG dataset, higher values of $\alpha$ lead to better performance, with $\alpha = 10.0$ achieving the lowest gap of $68.9\%$. However, on the C1 dataset, $\alpha = 0.1$ is optimal ($4.3\%$ gap), while $\alpha = 10.0$ causes severe degradation ($59.1\%$ gap). The C2 dataset also favors $\alpha = 0.1$ ($6.6\%$ gap), with performance degrading for higher values.

\begin{table}[h]
    \caption{Performance of \gls{mip} Random Perturbation with multiplicative noise ($\beta$). Results show the average gap (\%) across different datasets and parameter values.}
    \label{tab:mip_beta_results}
    \centering
    \setlength{\tabcolsep}{4pt}
    \begin{tabular}{lcccc}
        \toprule
        \textbf{Set} & \boldmath{$\beta = 1.1$} & \boldmath{$\beta = 1.5$} & \boldmath{$\beta = 2.0$} & \boldmath{$\beta = 5.0$} \\
        \midrule
        RG & $\mathbf{69.1 \pm 72.8}$ & $70.7 \pm 76.0$ & $70.3 \pm 76.3$ & $69.9 \pm 74.6$ \\
        C1 & $4.9 \pm 4.2$ & $4.8 \pm 3.6$ & $\mathbf{4.3 \pm 3.2}$ & $4.4 \pm 3.9$ \\
        C2 & $6.6 \pm 2.7$ & $\mathbf{6.4 \pm 1.9}$ & $7.5 \pm 3.2$ & $6.9 \pm 2.8$ \\
        \bottomrule
    \end{tabular}
\end{table}

For the multiplicative perturbation parameter $\beta$, the results show that parameters have slightly different effects on datasets, indicating that the parameters are not very relevant, especially under such tight \gls{mip} solving time limits. The RG dataset performs best with $\beta = 1.1$ ($69.1\%$ gap), while C1 favors $\beta = 2.0$ ($4.3\%$ gap) and C2 performs best with $\beta = 1.5$ ($6.4\%$ gap). However, $\alpha = 0.1$ and $\beta = 1.5$ are selected as the best parameters since they perform well on average across all datasets.

\subsection{MIP-based Biased Perturbation}

\begin{figure}[h]
    \centering
    \includegraphics[width=0.49\textwidth]{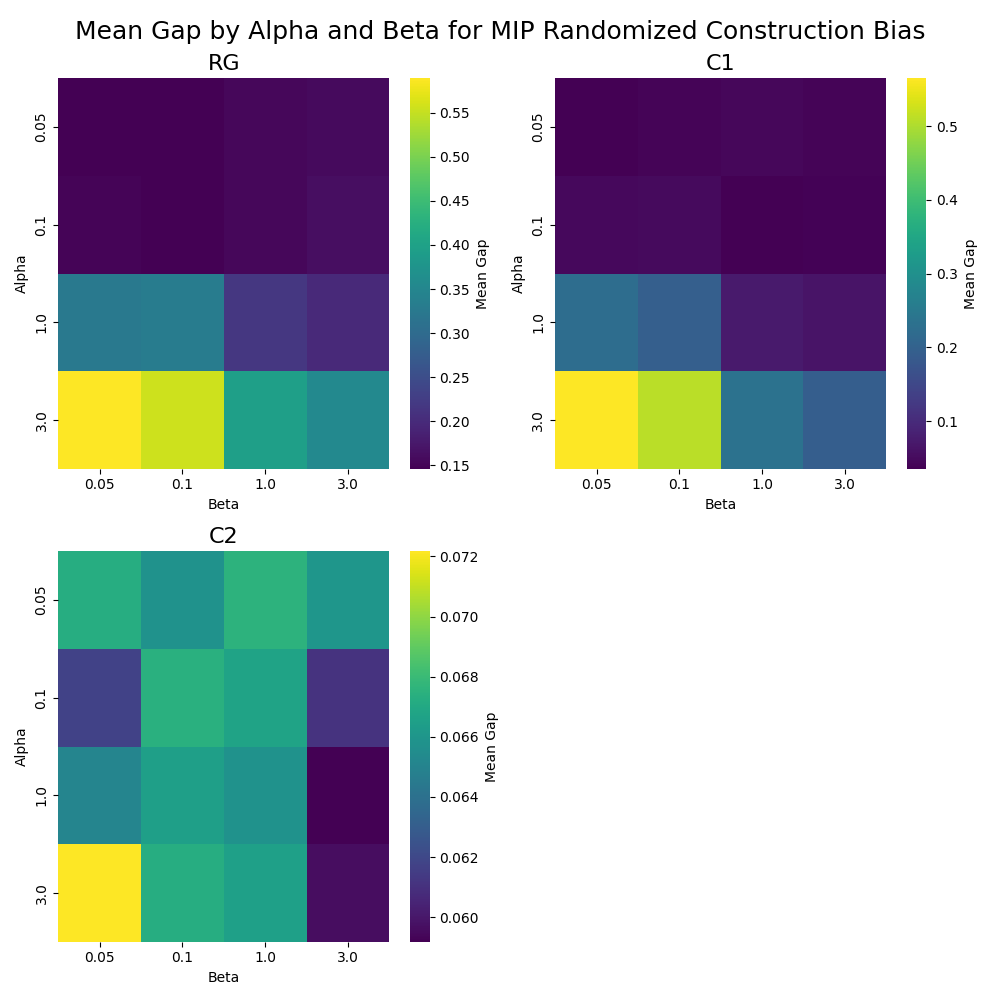}
    \caption{Performance comparison of \gls{mip}-based biased perturbation with parameters $\alpha$ and $\beta$ on different datasets. Results show the average gap (\%) across different parameter combinations.}
    \label{fig:alpha_beta_grid_mip_randomized_construction_bias} 
\end{figure}

The performance of the \gls{mip}-based biased perturbation heuristic is evaluated
using the grid of values $\alpha \in [0.05, 0.1, 1.0, 3.0]$ and $\beta \in [0.05, 0.1, 1.0, 3.0]$.
Figure~\ref{fig:alpha_beta_grid_mip_randomized_construction_bias} shows the heatmap of average gaps across different $\alpha$ and $\beta$ combinations for each dataset.
The darker colors indicate better performance (lower gaps), allowing identification of optimal parameter combinations for each dataset. 

For the RG and C1 datasets, higher values of $\alpha$ are better, while for the C2 dataset, lower values of $\alpha$ are better.
On all datasets, higher values of $\beta$ are better.
The following parameter values are used for the rest of the experiments: $\alpha = 0.1$ and $\beta = 3.0$, which are a good trade-off in all datasets.

\subsection{Comparison of Construction Heuristics}

A comparison of the performance of the tuned construction heuristics on all datasets is presented in Table~\ref{tab:construction_comparison}.
The Simple Randomized Construction, detailed in Section \ref{sec:constr_heuristic}, is referred to by the acronym SRC.

All datasets show the same ranking pattern: \gls{mip} biased construction consistently achieves the best performance, followed by \gls{mip} $\alpha$-Rand and \gls{mip} $\beta$-Rand, with RGC and SRC performing significantly worse.
The \gls{mip}-based methods demonstrate superior solution quality across all datasets, with gaps ranging from 3.8\% to 6.6\% on competition instances (C1, C2) and around 70\% on synthetic instances (RG).
This suggests that leveraging a TSP solver to establish a strong underlying tour structure is highly effective, even when the costs are perturbed.
In the case of the \gls{mip}-Bias method, the perturbations successfully guide the search towards regions of the solution space that are favorable for the TA-TSP objective, which provides evidence that introducing the trigger-target relations into the TSP model is effective.
In contrast, the simpler heuristics (SRC, RGC) show much higher gaps, particularly on the RG dataset where SRC reaches nearly 300\% gap. Notably, SRC fails completely on C2 (0\% success rate), while RGC shows poor performance on C2 (2.9\% success rate).
The \gls{mip} methods maintain 100\% success rates because finding feasibility for the TSP is easy on these datasets.
Runtime analysis reveals that SRC and RGC are the fastest (under 0.1 seconds), while \gls{mip}-based methods require 1-6 seconds.
However, all methods except for Gurobi are fast enough to be used as the construction heuristic backbone of a \gls{grasp} framework.
Given these results, the construction heuristic used in the \gls{grasp} framework is \gls{mip} biased construction.

\begin{table}[h]
    \caption{Performance comparison of construction heuristics across all datasets. Results show average gap (\%), runtime (seconds), and success rate for each method-dataset combination.}
    \label{tab:construction_comparison}
    \centering
    \setlength{\tabcolsep}{2.1pt}
    \begin{tabular}{llrrr}
        \toprule
        \textbf{Set} & \textbf{Method} & \textbf{Gap (\%)} & \textbf{Time (s)} & \textbf{Success} \\
        \midrule
        \multirow[c]{6}{*}{C1} & SRC & $163.6 \pm 128.5$ & $0.0 \pm 0.0$ & $0.95$ \\
        & RGC ($\alpha = 0.1$) & $29.2 \pm 24.6$ & $0.1 \pm 0.1$ & $0.91$ \\
        & \gls{mip} $\alpha$-Rand(0.1) & $4.3 \pm 3.2$ & $3.2 \pm 3.7$ & $1.00$ \\
        & \gls{mip} $\beta$-Rand(1.5) & $4.8 \pm 3.7$ & $3.2 \pm 3.4$ & $1.00$ \\
        & \gls{mip} Bias(0.1, 3.0) & $\mathbf{3.8 \pm 2.8}$ & $4.1 \pm 5.2$ & $1.00$ \\
        \midrule
        \multirow[c]{6}{*}{C2} & SRC & $-$ & $0.0 \pm 0.0$ & $0.00$ \\
        & RGC ($\alpha = 0.1$) & $2.7$ & $1.5 \pm 1.7$ & $0.03$ \\
        & \gls{mip} $\alpha$-Rand(0.1) & $6.6 \pm 2.2$ & $6.1 \pm 3.8$ & $1.00$ \\
        & \gls{mip} $\beta$-Rand(1.5) & $6.4 \pm 1.9$ & $5.9 \pm 3.7$ & $1.00$ \\
        & \gls{mip} Bias(0.1, 3.0) & $\mathbf{6.1 \pm 1.8}$ & $1.9 \pm 1.5$ & $1.00$ \\
        \midrule
        \multirow[c]{6}{*}{RG} & Gurobi & $85.1 \pm 192.5$ & $26.9 \pm 27.5$ & $1.00$ \\
        & SRC & $296.4 \pm 236.0$ & $0.0 \pm 0.0$ & $1.00$ \\
        & RGC ($\alpha = 0.1$) & $66.8 \pm 59.0$ & $0.0 \pm 0.0$ & $1.00$ \\
        & \gls{mip} $\alpha$-Rand(0.1) & $70.9 \pm 75.8$ & $1.1 \pm 1.9$ & $1.00$ \\
        & \gls{mip} $\beta$-Rand(1.5) & $70.7 \pm 76.2$ & $1.1 \pm 1.9$ & $1.00$ \\
        & \gls{mip} Bias(0.1, 3.0) & $\mathbf{69.7 \pm 75.1}$ & $1.1 \pm 1.8$ & $1.00$ \\
        \bottomrule
    \end{tabular}
\end{table}

\subsection{GRASP Framework}

The results obtained using the GRASP framework are presented in this section. Building upon the findings from the previous section, the MIP Bias construction heuristic is employed with parameters set to $\alpha = 0.1$ and $\beta = 3.0$. A time constraint of 60 seconds is allocated for the entire algorithm, with a specific limit of 2 seconds designated for the MIP solver. The GRASP framework operates without any restrictions on the number of trials or iterations.
The GRASP framework is evaluated using three local search operators: Relocate, SwapTwo, and TwoOpt. Additionally, an ablation study is conducted by sequentially removing each operator, and the performance of the GRASP framework is reported for each scenario.

\begin{table}[h]
    \caption{Performance of \gls{grasp} framework variants across different datasets. Results show the average gap (\%) for different local search configurations. The smallest (best) gap in each dataset is \textbf{bolded}.}
    \label{tab:grasp_performance}
    \centering
    \begin{tabular}{llr}
        \toprule
        \textbf{Set} & \textbf{Method} & \textbf{Gap (\%)} \\
        \midrule
        \multirow[c]{4}{*}{RG} 
            & \gls{grasp} & $-11.33 \pm 26.20$ \\
            & w/o Relocate & $-8.35 \pm 24.93$ \\
            & w/o SwapTwo & $-11.56 \pm 25.75$ \\
            & w/o TwoOpt & $\mathbf{-11.59 \pm 25.80}$ \\
        \midrule
        \multirow[c]{4}{*}{C1} 
            & \gls{grasp} & $\mathbf{0.77 \pm 1.22}$ \\
            & w/o Relocate & $4.85 \pm 10.02$ \\
            & w/o SwapTwo & $1.58 \pm 3.46$ \\
            & w/o TwoOpt & $2.72 \pm 6.86$ \\
        \midrule
        \multirow[c]{4}{*}{C2} 
            & \gls{grasp} & $\mathbf{0.40 \pm 0.41}$ \\
            & w/o Relocate & $0.99 \pm 0.74$ \\
            & w/o SwapTwo & $0.50 \pm 0.45$ \\
            & w/o TwoOpt & $\mathbf{0.40 \pm 0.35}$ \\
        \bottomrule
    \end{tabular}
\end{table}

The ablation results clearly indicate that using all three local search operators (Relocate, SwapTwo, and TwoOpt) simultaneously yields the best overall performance.
The Relocate operator is particularly important on all datasets, since removing it leads to a significant increase in the gap.
The \gls{grasp} framework is shown to be very effective in providing fast and high-quality solutions for the \gls{tatsp}.
The negative gaps on the RG dataset demonstrate that even on small instances, \gls{grasp} is more effective than the Gurobi \gls{mip} solver when both are operating on short time limits (60 seconds).
On the C1 and C2 datasets, the \gls{grasp} framework is able to provide final mean gaps of 0.77\% and 0.40\% respectively, which is an excellent performance for a heuristic running in a 1-minute time budget.

\paragraph{GRASP settings for the MESS 2024 competition}
\label{sec:grasp_settings_for_competition}

When solving the competition instances during the competition, the full \gls{grasp} with the \gls{mip} biased construction heuristic with parameters $\alpha = 0.1$ and $\beta = 3.0$ was used.
All three local search operators were used, as suggested by the ablation results.
The time limit of the \gls{mip} solver was set to 5 seconds, and the time limit of the \gls{grasp} to 3 hours per instance.
No limit was imposed on the number of trials or iterations of the \gls{grasp}.

The best solutions and the final objective values for each of the competition instances are provided in the repository.

\section{Conclusion}
This work presents a fast and effective GRASP-based metaheuristic for the recently introduced Trigger Arc Traveling Salesman Problem (TA-TSP). The approach integrates a novel MIP-based construction heuristic with a multi-neighborhood local search procedure to efficiently explore the complex, path-dependent solution space of the TA-TSP.

Computational experiments on diverse benchmark instances, including those from the MESS 2024 competition, demonstrate the method’s effectiveness. The GRASP framework consistently finds high-quality solutions within a one-minute time budget, significantly outperforming exact MIP solvers such as Gurobi, which often struggle to find feasible solutions or useful bounds for large-scale instances. This algorithm’s strong performance earned a top-three position in the MESS 2024 competition.

Future research could extend this work by exploring more sophisticated local search neighborhood structures tailored to the unique trigger-target dynamics of the \gls{tatsp}.
Furthermore, a more extensive hyperparameter tuning process, potentially using automated algorithm configuration tools, could further enhance the robustness and performance of the proposed \gls{grasp} algorithm.

\appendix
\section{MIP Model for the TA-TSP}
\label{sec:mip_model}
For benchmarking purposes, a \gls{mip} model for the \gls{tatsp} was developed, which shares similarities with the ILP model proposed by~\citet{Cerrone}.
The full model is presented in the Appendix.

The model uses the following parameters, sets, and decision variables:
\begin{itemize}
    \item $V$: Set of nodes $\{0, 1, \dots, N-1\}$.
    \item $E$: Set of directed arcs $(i,j)$.
    \item $c_{ij}$: Base cost of traversing arc $(i,j)$.
    \item $R_a$: Set of trigger arcs for a target arc $a \in E$.
    \item $r_{ba}$: Additional cost incurred if trigger arc $b$ activates target arc $a$.
    \item $x_{ij} \in \{0, 1\}$: 1 if arc $(i,j)$ is in the tour, 0 otherwise.
    \item $u_i \in [0, N-1]$: Position of node $i$ in the tour. For an arc $a=(i,j)$, $u_a$ denotes $u_i$.
    \item $y_{ba} \in \{0, 1\}$: 1 if relation $(b,a)$ is active, 0 otherwise.
    \item $z_{a_ia_j} \in \{0, 1\}$: 1 if arc $a_i$ precedes arc $a_j$ in the tour, 0 otherwise.
\end{itemize}

\footnotesize

The objective function minimizes the total tour cost:
\begin{equation} \label{eq:obj}
\min \sum_{(i,j) \in E} c_{ij} x_{ij} + \sum_{a \in E} \sum_{(b,a) \in R_a} r_{ba} y_{ba}
\end{equation}

Subject to the following constraints:
\begin{subequations}
\begin{align}
    \sum_{j \in V} x_{ij} &= 1 \quad \forall i \in V \label{eq:flow_out} \\
    \sum_{j \in V} x_{ji} &= 1 \quad \forall i \in V \label{eq:flow_in} \\
    u_i - u_j + N x_{ij} &\leq N-1 \quad \forall (i,j) \in E, j \neq 0 \label{eq:mtz} \\
    u_0 &= 0 \label{eq:start_node} \\
    \sum_{(b,a) \in R_a} y_{ba} &\leq x_a \quad \forall a \in E \label{eq:rel_active_1} \\
    y_{ba} &\leq x_b \quad \forall a \in E, (b,a) \in R_a \label{eq:rel_active_2} \\
    u_b + 1 &\leq u_a + N(1-y_{ba}) \quad \forall a \in E, (b,a) \in R_a \label{eq:rel_precedence} \\
    1 - z_{ab} &\leq \sum_{(c,a) \in R_a} y_{ca} + (1 - x_a) + (1 - x_b) \notag \\
    &\quad \forall a \in E, (b,a) \in R_a \label{eq:rel_active_3} \\
    u_{a_i} &\leq u_{a_j} + (N-1)(1-z_{a_ia_j}) \quad \forall a_i, a_j \in E \label{eq:z_precedence} \\
    y_{ba} &\leq y_{ca} + z_{cb} + z_{ac} + (1-x_c) + (1-x_b) + (1-x_a) \notag \\
    &\quad \forall a \in E, (b,a),(c,a) \in R_a, b \neq c \label{eq:rel_last_trigger}
\end{align}
\end{subequations}

\normalsize

The objective function~\eqref{eq:obj} minimizes the total tour cost, composed of the base arc costs and the additional costs from active relations.

Constraints~\eqref{eq:flow_out} and~\eqref{eq:flow_in} are flow conservation constraints ensuring that each node is visited exactly once.
Constraints~\eqref{eq:mtz} and~\eqref{eq:start_node} are the Miller-Tucker-Zemlin (MTZ) formulations for subtour elimination, which define the sequence of nodes in the tour, with node 0 as the starting point.

The relation activation logic is modeled by constraints~\eqref{eq:rel_active_1} through~\eqref{eq:rel_last_trigger}.
Constraint~\eqref{eq:rel_active_1} ensures that a relation for a target arc $a$ can only be active if $a$ is part of the tour, and at most one such relation can be active.
Constraint~\eqref{eq:rel_active_2} strengthens this by requiring the trigger arc $b$ to also be in the tour for the relation $(b,a)$ to be active.
Constraint~\eqref{eq:rel_precedence} states that for a relation $(b,a)$ to be active, the trigger arc $b$ must be traversed before the target arc $a$.
The precedence variable $z_{a_ia_j}$ is defined in constraint~\eqref{eq:z_precedence}, where $z_{a_ia_j}=1$ if arc $a_i$ precedes arc $a_j$.
Constraint~\eqref{eq:rel_active_3} ensures that if an arc $a$ and one of its potential triggers $b$ are in the tour, and $b$ precedes $a$, then at least one relation for $a$ must be activated.
Finally, constraint~\eqref{eq:rel_last_trigger} is the core constraint that models the "last trigger" rule. It ensures that if two triggers $b$ and $c$ for the same target $a$ are in the tour, and $c$ is traversed after $b$ but before $a$, then the relation $(b,a)$ cannot be active.

%%%%%%%%%%%%%%   Bibliography   %%%%%%%%%%%%%%
\normalsize
\bibliography{references}

\end{document}